\def\year{2021}\relax
\documentclass[letterpaper]{article} 
\usepackage{aaai21}  
\usepackage{times}  
\usepackage{helvet} 
\usepackage{courier}  
\usepackage[hyphens]{url}  
\usepackage{graphicx} 
\usepackage{natbib}  
\usepackage{caption} 
\usepackage{amsmath,amssymb}
\usepackage{subfigure}
\usepackage[switch]{lineno}  %
\usepackage{xcolor}
\newcommand{\att}[1]{\text{\tiny{\textbf{#1}}}}
\title{Dynamic Graph Representation Learning for Video Dialog \\via Multi-Modal Shuffled Transformers}

\author{Shijie Geng$^{\dag}$, Peng Gao$^{\ddag}$, Moitreya Chatterjee$^{\S}$, Chiori Hori$^\diamond$,\\
Jonathan Le Roux$^\diamond$, Yongfeng Zhang$^\dag$, Hongsheng Li$^\ddag$, Anoop Cherian$^\diamond$\\}

\affiliations{
    $^\dag$Department of Computer Science, Rutgers University, Piscataway, NJ, USA \\
    $^\ddag$The Chinese University of Hong Kong \quad $^\S$University of Illinois at Urbana Champaign, Urbana, IL, USA\\
    $^\diamond$Mitsubishi Electric Research Laboratories (MERL), Cambridge, MA, USA \\
}

\begin{document}
\maketitle

\begin{abstract}
Given an input video, its associated audio, and a brief caption, the audio-visual scene aware dialog (AVSD) task requires an agent to indulge in a question-answer dialog with a human about the audio-visual content. This task thus poses a challenging multi-modal representation learning and reasoning scenario, advancements into which could influence several human-machine interaction applications. To solve this task, we introduce a \emph{semantics-controlled multi-modal shuffled Transformer reasoning} framework, consisting of a sequence of Transformer modules, each taking a modality as input and producing representations conditioned on the input question. Our proposed Transformer variant uses a shuffling scheme on their multi-head outputs, demonstrating better regularization. To encode fine-grained visual information, we present a novel dynamic scene graph representation learning pipeline that consists of an \emph{intra-frame reasoning} layer producing spatio-semantic graph representations for every frame, and an \emph{inter-frame aggregation} module capturing temporal cues. Our entire pipeline is trained end-to-end. We present experiments on the benchmark AVSD dataset, both on answer generation and selection tasks. Our results demonstrate state-of-the-art performances on all evaluation metrics.
\end{abstract}
\section{Introduction}

\begin{figure}[t]
  \centering
  \includegraphics[width=12cm,trim={1.2cm 8.9cm 2.5cm 0.5cm},clip]{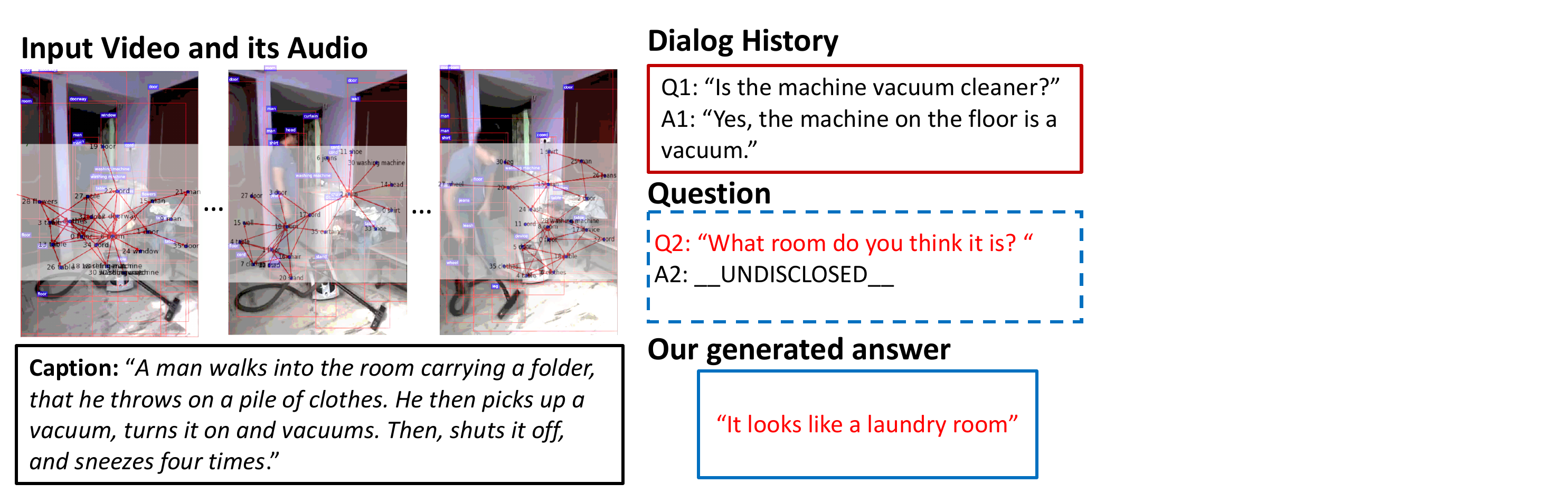}
\caption{A result from our proposed model for the AVSD task. Given a video clip, its caption,  dialog history, and a question, the AVSD generation task aims to generate the answer in natural language form.}
\label{fig:first_page}
\vspace{-5pt}
\end{figure}

The success of deep learning in producing effective solutions to several fundamental problems in computer vision,  natural language processing, and speech/audio understanding has provided an impetus to explore more complex multi-modal problems at the intersections of these domains, attracting wide interest recently~\cite{zhu2020deep}. A few notable ones include: (i) visual question answering (VQA)~\cite{antol2015vqa,yang2003videoqa}, the goal of which is to build an agent that can generate correct answers to free-form questions about visual content, (ii) audio/visual captioning~\cite{hori2017attention,venugopalan2015sequence,xu2015show,drossos2019clotho}, in which the agent needs to generate a sentence in natural language describing the audio/visual content, (iii) visual dialog~\cite{das2017visual}, in which the agent needs to engage in a natural conversation with a human about a static image, and (iv) audio-visual scene-aware dialog (AVSD)~\cite{alamri2019audio,hori2019end} -- that generalizes (i), (ii), and (iii) -- in which the agent needs to produce a natural answer to a question about a given audio-visual clip,  in a conversation setting or select the correct answer from a set of candidates. The AVSD task\footnote{\url{https://video-dialog.com/}} emulates a real-world human-machine conversation setting that is potentially useful in a variety of practical applications, such as building virtual assistants~\cite{deruyttere2019talk2car} or in human-robot interactions~\cite{thomason2019improving}. See Figure~\ref{fig:first_page} for an illustration of this task.

The generality of the AVSD task, however, poses a challenging multi-modal representation learning and reasoning problem.  Specifically, some of the input modalities to this task may offer complementary information (such as video and audio), while a few others may be independent (audio and captions), or even conflict with each other, e.g., the provided text (captions/dialogs) may include details from human experience that are absent in the video (e.g., ``I think...''), or may include abstract responses (``happy'', ``bored'', etc.) that may be subjective. Thus, the main quest in this task is to represent these modalities such that inference on them is efficient and effective. Previous approaches to this problem  used holistic video features produced by a generic 3D convolutional neural network~\cite{carreira2017quo}, and either focused on extending attention models on these features to include additional modalities~\cite{alamri2019audio,hori2019end,schwartz2019simple}, or use vanilla Transformer networks~\cite{vaswani2017attention} to produce effective multi-modal embeddings~\cite{le2019multimodal}. These off-the-shelf visual representations or Transformer architectures are not attuned to the task, potentially leading to sub-optimal performance.

In this paper, we present a neural inference algorithm that hierarchically reduces the complexity of the AVSD task using the machinery of graph neural networks and sequential multi-head Transformers. Specifically, we first present a spatio-temoral scene graph representation (STSGR) for encoding the video compactly while capturing its semantics. Specifically, our scheme builds on visual scene graphs~\cite{johnson2015image} towards video representation learning by introducing two novel modules: (i) an intra-frame reasoning module that combines graph-attention~\cite{velivckovic2017graph} and edge-convolutions~\cite{wang2019dynamic} to produce a semantic visual representation for every frame, (ii) subsequently, an inter-frame aggregation module uses these representations and updates them using information from temporal-neighborhoods, thereby producing compact spatio-temporal visual memories. We then couple these memories with temporally aligned audio features. Next, multi-head Transformers~\cite{vaswani2017attention}, encodes each of the other data modalities (dialog history, captions, and the pertinent question) separately alongside these audio-visual memories and fuses them sequentially using Transformer decoders. These fused features are then used to select or generate the \emph{answers} auto-regressively. We also present a novel extension of the standard multi-head Transformer network in which the outputs of the heads are mixed. We call this variant a \emph{shuffled Transformer}. Such random shuffling avoids overfitting of the heads to its inputs, thereby regularizing them, leading to better generalization.

To empirically evaluate our architecture, we present experiments on two variants of the AVSD dataset available as part of the 7th and 8th Dialog System Technology Challenges (DSTC). We provide experiments on both the answer generation and the answer selection tasks -- the former requiring the algorithm to produce free-form sentences as answers, while the latter selects an answer from 100 choices for each question. Our results reveal that using the proposed STSGR and our shuffled Transformer lead to significant improvements on both tasks against state-of-the-art methods on all metrics.
The key contributions of this paper are:
\begin{itemize}
\itemsep0em
\item We propose to represent videos as spatio-temporal scene graphs capturing key audio-visual cues and semantic structure. To the best of our knowledge, the combination of our intra/inter-frame reasoning modules is novel. 
\item We introduce a sequential Transformer architecture that uses shuffled multi-head attention, yielding question-aware representations of each modality while generating answers (or their embeddings) auto-regressively.
\item  Extensive experiments on the AVSD answer generation and selection tasks demonstrate the superiority of our approach over several challenging recent baselines.
\end{itemize}

\begin{figure*}[t!]
  \centering
  \includegraphics[width=16.6cm, height=6.7cm]{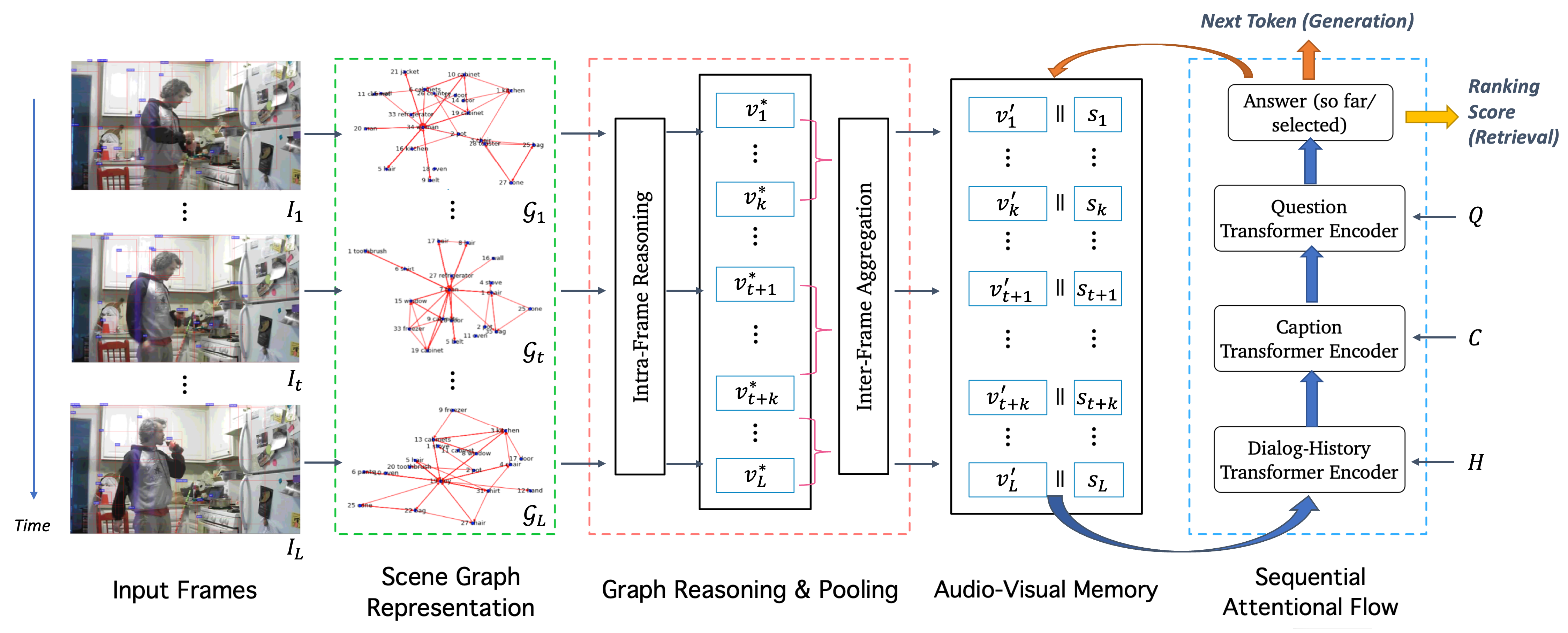}
\caption{A schematic illustration of our overall pipeline for dialog response generation/retrieval.}
\label{fig:model}
\vspace*{-.35cm}
\end{figure*}
\section{Related Work}
Our proposed framework has similarities with prior works along three different axes, viz. (i) graph-based reasoning, (ii) multi-modal attention, and (iii) visual dialog methods. 

\noindent \textbf{Scene Graphs:}~\cite{johnson2015image} combine objects detected in static images, their attributes, and object-object relationships~\cite{lu2016visual} to form a directed graph that not only provides an explicit and interpretable representation of the image, but is also seen to be beneficial for higher-order reasoning tasks such as image captioning~\cite{li2019know,yang2019auto}, and visual question answering~\cite{ghosh2019generating,norcliffe2018learning,geng20192nd,geng2020character}. There have been efforts~\cite{wang2018non,girdhar2019video,jain2016structural,herzig2019spatio,jang2017tgif,tsai2019video} at capturing spatio-temporal evolution of localized visual cues. In~\cite{Wang2018videos}, a space-time graph reasoning framework is proposed for action recognition.  Similarly, the efficacy of manually-labeled video scene graphs is explored in~\cite{ji2019action}. Similar to ours, they use object detections per video frame, and construct a spatio-temporal graph based on the affinities of the features from the detected objects. Spatio-temporal graphs using knowledge distillation is explored for video captioning in~\cite{pan2020spatio}. In contrast, our task involves several diverse modalities, demanding richer architectural choices. Specifically, we present a hierarchically organized intra/inter-frame reasoning pipeline for generating visual memories, trained via neural message passing, offering a powerful inference engine. Our ablation studies demonstrate the usefulness of these modules. 
 
\noindent\textbf{Multi-modal Fusion/Attention:}  has been explored in several prior works~\cite{hori2017attention,hori2018multimodal,hori2019end,shi2020multi}, however does not use the power of Transformers. Self-attention and feature embeddings using Transformers is attempted in multi-modal settings~\cite{Gao_2019_CVPR,Gao_2019_ICCV,shi2020contrastive}, however only on static images. Bilinear fusion methods~\cite{ben2019block,fukui2016multimodal} have been explored towards inter-modality semantic alignment, however they often result in high-dimensional interaction tensors that are computationally expensive during inference. In contrast, our pipeline is the first to leverage the power of Transformers in a hierarchical graph reasoning setup for video dialogs and is cheap to compute.
 
\noindent\textbf{Multi-modal Dialogs:} have been explored in various ways before. Free-form human-like answers were first considered in~\cite{das2017visual}, which also proposed the VisDial dataset, however on static images. A difficulty in designing algorithms on multi-modal data is in deriving effective attention mechanisms that can divulge information from disparate cues. To tackle this challenge,~\cite{wu2018you} proposed a sequential co-attention scheme in which the neural embeddings of various modalities are co-attended with visual embeddings in a specific order. ~\cite{schwartz2019factor} generalized the co-attention problem by treating the modalities as nodes of a graph, aggregating them as \emph{factors}, using neural message passing. We use a combination of these two approaches; specifically we use Transformer encoders for embedding each modality, and attend on these multi-modal embeddings sequentially to generate the answer. Further, in contrast to~\cite{schwartz2019factor,wu2018you}, that tackle solely the answer generation problem, we consider the answer selection task on AVSD as well. ~\cite{yeh2019reactive} also proposed using Transformers~\cite{vaswani2017attention} for fusing audio-visual features on the AVSD task. A multi-step reasoning scheme is proposed in~\cite{gan2019multi} using joint attention via an RNN for generating a multi-modal representation. The Simple baseline~\cite{schwartz2019simple} extends factor graphs~\cite{schwartz2019factor} for the AVSD problem demonstrating promising results. A multi-modal Transformer for embedding various modalities and a query-aware attention is introduced in~\cite{le2019multimodal}. ~\cite{le2020video} fine-tunes pretrained GPT-2 to obtain improved performance. However, these works neither consider richer visual representations using scene graphs, nor variations of the standard Transformers, like the shuffling scheme we present.
\section{Proposed Method}
In this section, we will first present our spatio-temporal scene graph representation (STSGR) for encoding the video sequences, following which we elaborate on our multi-modal shuffled Transformer architecture.

\subsection{Overview of Spatio-Temporal Scene Graphs}
Given a video sequence $V$, let $C$ denote the associated human-generated video caption, and let $(Q_i, A_i)$ represent the tuple of the text-based $i$-th question and its answer in the given human dialog about $V$ (see Fig.~\ref{fig:first_page}). We will succinctly represent the dialog history by $H = \langle(Q_1, A_1),\dots, (Q_{l-1}, A_{l-1})\rangle$. Further, let $Q_l$ represent the question under consideration. The audio-visual scene-aware dialog (AVSD) task requires the generation (or selection) of the answer denoted by $A_l$, corresponding to the question $Q_l$. 

Our proposed pipeline to solve this task is schematically illustrated in Fig.~\ref{fig:model}. It consists of four components: (1) a \emph{scene graph construction module}, which extracts objects and relation proposals from the video using pretrained neural network models, building a scene graph for every (temporally-sampled) video frame, (2) an \emph{intra-frame reasoning module}, which conducts node-level and edge-level graph reasoning, producing compact feature representations for each scene graph, (3) an \emph{inter-frame information aggregation module}, that aggregates these features within a temporal sliding window to produce a \emph{visual memory} for each frame's scene graph (at the center frame in that window), and (4) a \emph{semantics-controlled Transformer reasoning module}, which performs multi-modal reasoning and language modelling based on a semantic controller. In this module, we also use a newly-proposed shuffle-augmented co-attention to enable head interactions in order to boost performance. Below, we describe in detail each of these modules.

\subsection{Scene Graph Representation of Video}
Our approach to generate scene graphs for the video frames is loosely similar to the ones adopted in recent works such as~\cite{pan2020spatio,herzig2019spatio,Wang2018videos}, and has three components: (a) object detection, (b) visual-relation detection, and (c) region of interest (ROI) recrop on union bounding boxes. For (a), we train a Faster R-CNN model~\cite{ren2015faster} on the Visual Genome dataset~\cite{krishna2017visual} using the MMDetection repository \cite{mmdetection}. For a video frame $I$, this Faster-RCNN model produces: $\mathcal{F}_{I}, \mathcal{B}_I, \mathcal{S}_I = \mathrm{RCNN}(I)$, 
where $\mathcal{F}_{I} \in {\mathbb{R}^{N_o \times d_o}}$ denotes the $d_o$-dimensional object features, $\mathcal{B}_I  \in {\mathbb{R}^{N_o \times 4}}$ are the object bounding boxes, and $\mathcal{S}_{I}$ is a list of semantic labels associated with each bounding box. The pair $(\mathcal{F}_{I}, \mathcal{S}_I)$ forms the nodes of our scene graph. Next, to find the graph edges, we train a relation model on the VG200 dataset~\cite{krishna2017visual}, which contains 150 objects and 50 predicates, and apply this learned model on the frames from the given video. The output of this model is a set of $\langle S, P, O \rangle$ triplets per frame, where $S$, $P$, and $O$ represent the \textit{subject}, \textit{predicate}, and \textit{object}, respectively. We keep the $\langle S,O \rangle$ pairs as relation proposals and discard the original predicate semantics, as the relation predicates of the model trained on VG200 are limited and fixed. Instead, we let our reasoning model learn implicit relation semantics during our end-to-end training. For the detected $\langle S,O \rangle$ pairs, we regard the union box of the bounding boxes for $S$ and $O$ as the predicate region of interest. Next, we apply the \emph{ROI-align} operator~\cite{ren2015faster} on the last layer of the backbone network using this union box and make the resultant feature an extra node in the scene graph. 

\subsection{Intra-Frame Reasoning}
Representing videos directly as sequences of scene graphs leads to a complex graph reasoning problem that can be computationally challenging. To avoid this issue, we propose to hierarchically reduce this complexity by embedding these graphs into learned representation spaces. Specifically, we propose an intra-frame reasoning scheme that bifurcates a scene graph into two streams: (i) a \emph{visual scene graph} that generates a representation summarizing the visual cues captured in the graph nodes, and (ii) a \emph{semantic scene graph} that summarizes the graph edges. Formally, let us define a scene graph as $\mathcal{G} = \{(x_i, e_{ij}, x_j) \mid x_i, x_j \in \mathcal{V}, e_{ij} \in \mathcal{E} \}$, where $\mathcal{V}$ denotes the set of nodes consisting of single objects and $\mathcal{E}$ is the set of edges consisting of relations linking two objects. The triplet $(x_i, e_{ij}, x_j)$ indicates that the subject node $x_i$ and the object node $x_j$ are connected by the directed relation edge $e_{ij}$.  We denote by $\mathcal{G}_v$ and $\mathcal{G}_s$ the visual scene graph and the semantic scene graph respectively: the former is a graph attention network \cite{velivckovic2017graph} which computes an attention coefficient for each edge and updates node features based on these coefficients; the latter is based on EdgeConv \cite{wang2019dynamic}, which computes extra edge features based on node features and then updates the node features by aggregating the edge features linked to a given node. Both networks are explained in detail next. We combine these two complementary graph neural networks in a cascade to conduct intra-frame reasoning.

\noindent \textbf{Visual Scene Graph Reasoning:} For $M$ node features $\mathbf{X} = \{\mathbf{x}_1, \mathbf{x}_2,\dots, \mathbf{x}_M\}$ in a scene graph, multi-head self-attention~\cite{vaswani2017attention} is first performed for each pair of linked nodes. In each head $k$, for two linked nodes $\mathbf{x}_i$ and $\mathbf{x}_j$, the attention coefficient $\alpha^k_{ij}$ indicating the importance of node $j$ to node $i$ is computed by
\begin{equation}
    \alpha^k_{ij} = \frac{\exp \left(\sigma\left(\mathbf{\Theta}^\top_k [\mathbf{W}^k_{\!1} \mathbf{x}_{i} \parallel \mathbf{W}^k_{\!1} \mathbf{x}_{j}]\right)\right)}{\sum_{k \in \mathcal{N}_i}\exp \left(\sigma\left(\mathbf{\Theta}^\top_k [\mathbf{W}^k_{\!1} \mathbf{x}_{i} \parallel \mathbf{W}^k_{\!1} \mathbf{x}_{k}]\right)\right)},
    \label{eq:1}
\end{equation}
where $\parallel$ denotes feature concatenation, $\sigma$ is a nonlinearity (Leaky ReLU), $\mathcal{N}_i$ indicates the neighboring graph nodes of object $i$ (including $i$), $\mathbf{W}^k_{\!1} \in \mathbb{R}^{d_{h} \times d_\text{in}}$ is a (learned) weight matrix transforming the original features to a shared latent space, and $\mathbf{\Theta}_k \in \mathbb{R}^{2d_{h}}$ is the (learned)  attention weight vector. 
Using the attention weights $\alpha^k$ and a set of learned weight matrices $\mathbf{W}^k_{\!2}\in \mathbb{R}^{d_{h}/K \times d_\text{in}}$, we update the node features as:
\begin{equation}
    \mathbf{x}'_{i} = \big\lVert_{k=1}^{K} \sigma \Big (  \sum_{j \in \mathcal{N}_i} \alpha_{ij}^k \mathbf{W}_{\!2}^{k} \mathbf{x}_{j}  \Big ).
\end{equation}
Outputs of the $K$ heads are concatenated to produce $\mathbf{x}'_{i}\in\mathbb{R}^{d_h}$, which is used as input to the semantic graph network.

\noindent \textbf{Semantic Scene Graph Reasoning: } This sub-module captures higher-order semantics between nodes in the scene graph. To this end, EdgeConv~\cite{wang2019dynamic}, which is a multi-layer fully-connected network $h_{\mathbf{\Lambda}}$, is employed to generate edge features $\mathbf{e}_{ij}$ from its two connected node features $(\mathbf{x}'_i, \mathbf{x}'_j)$: $
    \mathbf{e}_{ij} = h_{\mathbf{\Lambda}}(\mathbf{x}'_i, \mathbf{x}'_j)$,
where $h_{\mathbf{\Lambda}}: \mathbb{R}^{d_h} \times \mathbb{R}^{d_h} \rightarrow \mathbb{R}^{d_h}$ is a nonlinear transformation with learnable parameters $\mathbf{\Lambda}$. We then obtain the output node features $\mathbf{x}^{\star}_i$ by aggregating features from the edges that are directed to the object node $i$, i.e.,
\begin{equation}
    \mathbf{x}^{\star}_i = \max_{j:(j,i) \in \mathcal{E}_i} \mathbf{e}_{ji},
    \label{eq:2}
\end{equation}
where $\mathcal{E}_i$ denotes the set of edges directed to node $i$. All object features inside the scene graph are updated by the above intra-frame feature aggregation.

\noindent \textbf{Memory Generation with Graph Pooling:}  After conducting intra-frame reasoning to obtain higher-level features for each node, we pool the updated graph into a memory for further temporal aggregation. Since different frame-level scene graphs have different numbers of nodes and edges, we adopt graph average pooling ($\mathrm{GAP}$) and graph max pooling ($\mathrm{GMP}$) \cite{lee2019self} to generate two graph memories and concatenate them to produce $V^{\star}$: 
\begin{equation}
    V^{\star} = \mathrm{GAP}(\mathbf{X}^{\star}, \mathcal{E}) \parallel \mathrm{GMP}(\mathbf{X}^{\star}, \mathcal{E}),
    \label{eq:3}
\end{equation}
where $\mathcal{E}$ denotes the scene graph connection structure, and $\mathbf{X}^{\star}$ the $M$ node features $\{\mathbf{x}^{\star}_1, \mathbf{x}^{\star}_2,\dots, \mathbf{x}^{\star}_M\}$ from~\eqref{eq:2}.

\subsection{Inter-Frame Information Aggregation}
Apart from the spatial graph representations described above, there is a temporal continuity of visual cues in the video frames that needs to be captured as well. To this end, we propose an inter-frame aggregation scheme that operates on the spatial graph embeddings. Specifically, for a sequence of scene graph memories $\langle v^{\star}_1, v^{\star}_2,\dots, v^{\star}_L\rangle$ of length $L$ produced using~\eqref{eq:3} on a sequence of $L$ frames, we use temporal sliding windows of size $\tau$ to update the graph memory of the center frame in each window by aggregating the graph memories of its neighboring frames in that window, both in the past and in the future. Let $F \in \mathbb{R}^{2d_{h} \times \tau}$ denotes a matrix of graph embeddings within this window of length $\tau$, then we perform window-level summarization over all frame memories within $F$ as: $\beta = \mathrm{softmax}(\Gamma^\top\tanh(\mathbf{W}_{\!\tau} F))$, where $\mathbf{W}_{\!\tau} \in \mathbb{R}^{2d_h \times 2d_h}$ is a learned weight matrix, $\Gamma \in \mathbb{R}^{2d_h}$ is a weight vector, and $\beta$ denotes the attention weights. We then use $\beta$ to update the memory $v_c$ of the center frame (in this window) by aggregating information across this window, as: $v'_c = F\beta^\top$. 
Repeating this step for all sliding windows, we get the final visual graph memory $V'=\left\langle v'_1, v'_2,\dots, v'_L\right\rangle$ aggregating both spatial and temporal information. We also augment these visual features with their temporally-aligned audio embeddings $\left\langle s_1, s_2, \cdots, s_L\right\rangle$ produced using an AudioSet VGGish network~\cite{hershey2017cnn}. 

\subsection{Semantics-Controlled Transformer Reasoning}
\label{sec:Transformer}
\begin{figure}[t]
    \centering
    \includegraphics[width=9.1cm,trim={0cm 0cm 0cm 0cm},clip]{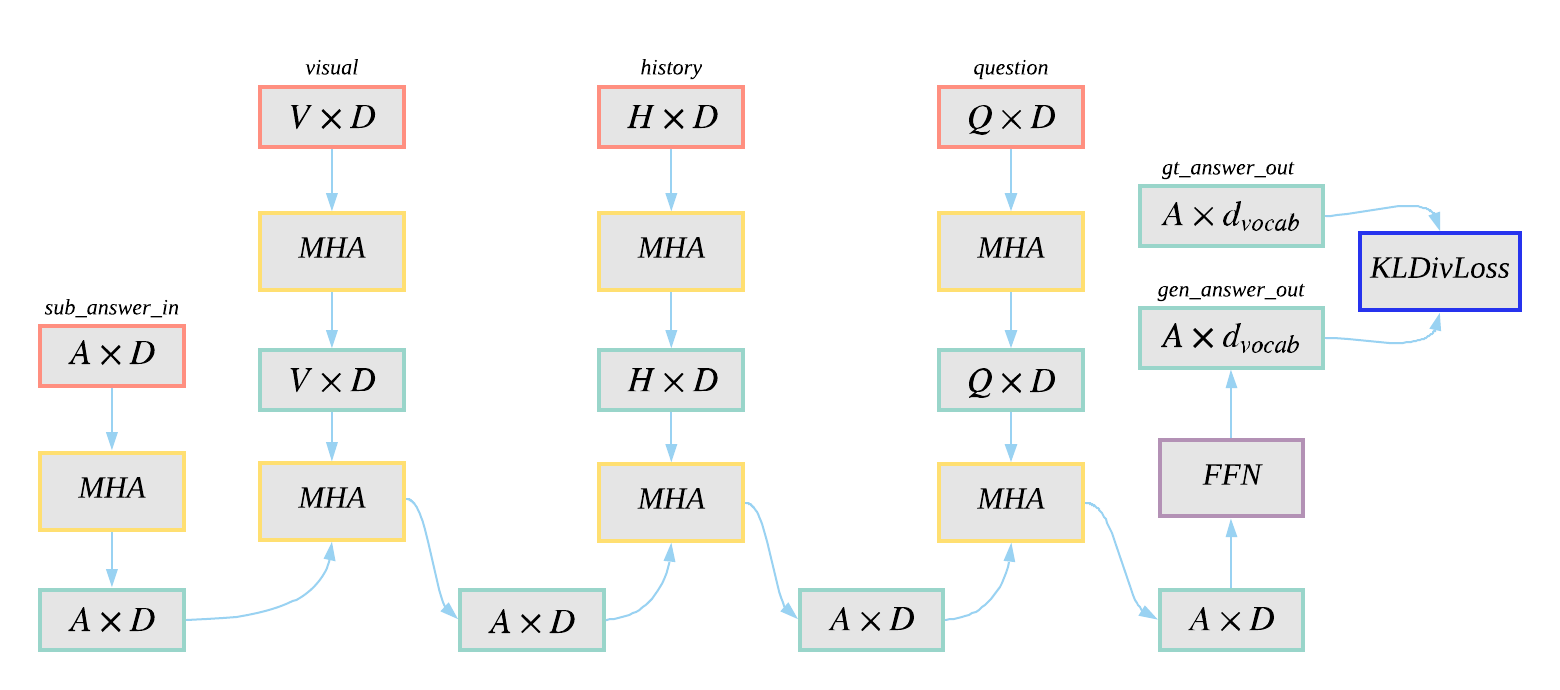}
    \caption{depicts the sequential attention flow in our semantics-controlled Transformer. MHA stands for multi-head attention. FFN is short for feed-forward networks. The acronyms A, V, H, and Q stand for the answers, visual memory, caption/dialog history, and the question, respectively. }
    \label{fig:Transformer}
    \vspace{-0.5cm}
\end{figure}

The above modules encode a video into a sequence of graph memories via reasoning on visual and semantic scene graphs. Besides encoding audio-visual information, we also need to encode the text information available in the AVSD task. For the sentence generation task, we propose to generate the answer autoregressively~\cite{anderson2018bottom,hori2018multimodal}, i.e., predict the next word in the answer from the vocabulary based on source sequences including the visual memory, query $Q_l$, caption $C$, the dialog history $H = \langle(Q_1, A_1),\dots, (Q_{l-1}, A_{l-1})\rangle$, and the partially generated answer so far, denoted $A_l^\text{in}$ (see  Fig.~\ref{fig:model} and Fig.~\ref{fig:Transformer}). This sub-answer $A_l^\text{in}$ forms the semantics that control the attention on the various modalities to generate the next word. As shown in Fig.~\ref{fig:Transformer}, our semantics-controlled Transformer module consists of a graph encoder, a text encoder, and a multi-modal decoder. It takes in source sequences and outputs the probability distribution of the next token for all tokens in the vocabulary. We detail the steps in this module next.

\noindent \textbf{Encoder:} We first use Transformer to embed all text sources ($H$, $C$, $Q_l$, $A_l^\text{in}$) using token and positional embeddings, generating feature matrices $e_h, e_c, e_{q}$, and $e_a$, each  of the same feature dimensionality $d_h$. We also use a single-layer fully-connected network to transfer the audio-augmented visual memories in $V'$ to $d_h$-dimensional features $e_v$ that match the dimension of the text sources. Next, for the answer generation task, the input sub-answer (generated so far) $e_a$ is encoded with a Transformer consisting of multi-head self-attention to get hidden representations $h^a_\text{enc}$:
\begin{equation}
    h^a_\text{enc} = \mathrm{FFN}^a(\mathrm{Attention}(\mathbf{W}^a_{\att{Q}} e_a, \mathbf{W}^a_\att{K} e_a, \mathbf{W}^a_\att{V} e_a)),
    \label{eq:6}
\end{equation}
\noindent where $\mathbf{W}^a_{\att{Q}}$, $\mathbf{W}^a_{\att{K}}$, $\mathbf{W}^a_{\att{V}}$ are weight matrices for query, key, and value respectively~\cite{vaswani2017attention}, $\mathrm{FFN}^a$ is a feed-forward module consisting of two fully-connected layers with ReLU in between. The $\mathrm{Attention}$ function is defined as in~\cite{vaswani2017attention}:
\begin{equation}
    \mathrm{Attention}(\mathbf{Q},~\mathbf{K},~\mathbf{V}) = \mathrm{softmax}(\frac{\mathbf{Q}\mathbf{K}^\top}{\sqrt{d_h}})\mathbf{V},
\end{equation}
\noindent with a scaling factor $\sqrt{d_h}$ that maintains the order of magnitude in features, and $\mathbf{Q},\mathbf{K},\mathbf{V}$ represent the query, key, and value triplets as described in~\cite{vaswani2017attention}. After encoding the input sub-answer, we conduct co-attention in turn for each of the other text and visual embeddings $e_j$, where $j \in \{v, c, h, q\}$, with a similar Transformer architecture. That is, the encoding $h^j_{\text{enc}}$ for a given embedding type $e_j$ is obtained by using the encoding $h^{j'}_{\text{enc}}$ for the previous embedding type $e_{j'}$ as query (Fig.~\ref{fig:Transformer}):
\begin{equation}
    h^j_{\text{enc}} = \mathrm{FFN}^j(\mathrm{Attention}(\mathbf{W}^j_{\att{Q}} h^{j'}_\text{enc}, \mathbf{W}^j_{\att{K}} e_j, \mathbf{W}^j_{\att{V}} e_j)).
\end{equation}
In our implementation, the embeddings for history and caption are concatenated as $e_{c+h}=e_c||e_h$. Processing occurs in the following order: starting from $h^a_\text{enc}$, we compute $h^v_{\text{enc}}$, then $h^{c+h}_{\text{enc}}$, and later $h^q_{\text{enc}}$.
Finally, we get a feature vector $h^{\star}_\text{enc}$ that fuses all the information from the text and audio-visual sources by concatenating these multi-modal features.

\begin{figure}[t]
    \centering
    \includegraphics[width=7.5cm,trim={0cm 0cm 0cm 0cm},clip]{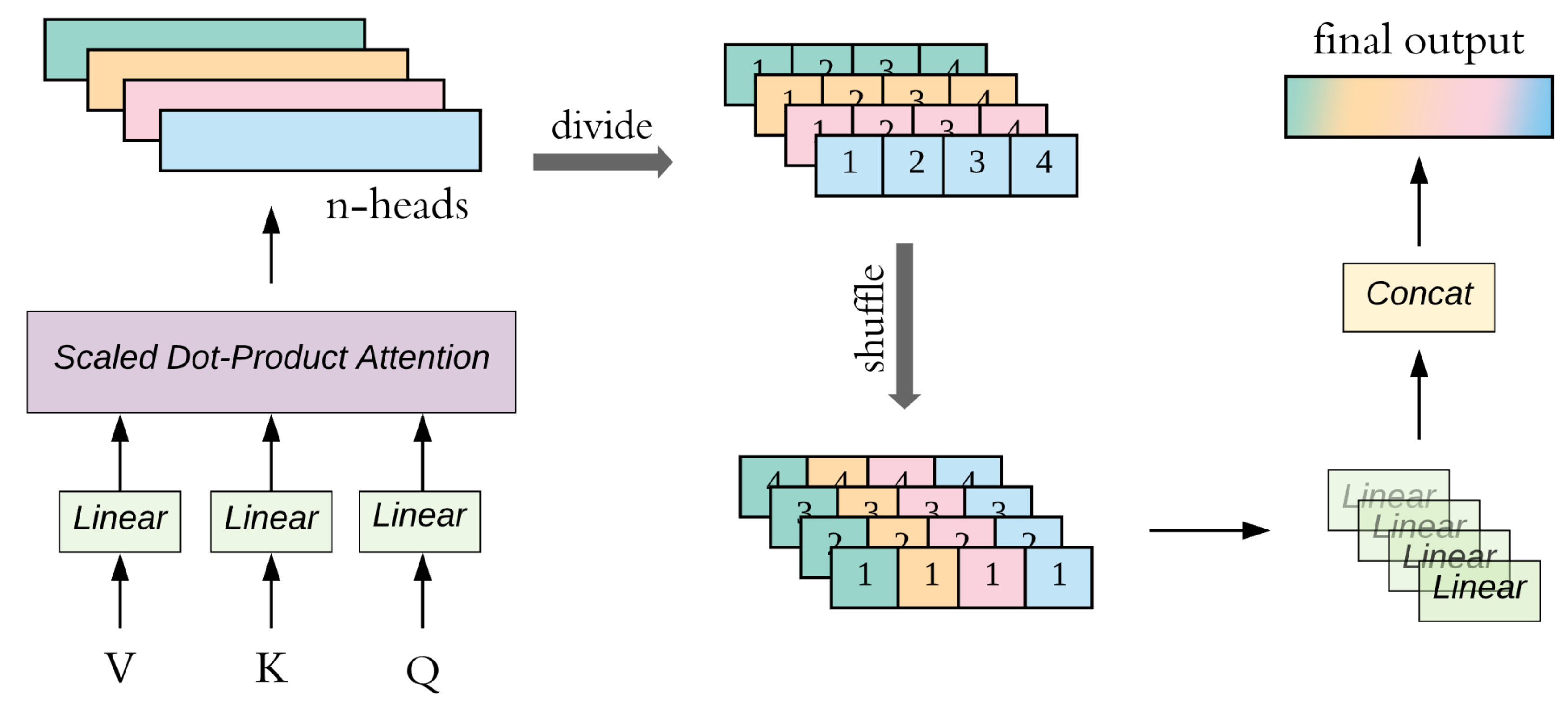}
    \vspace{-0.2cm}
    \caption{An illustration of our multi-head shuffled Transformer, where we shuffle the output of each head before passing it on to the FFN module.}
    \label{fig:shuffle-txr}
    \vspace{-0.55cm}
\end{figure}

\noindent\textbf{Multi-head Shuffled Transformer:} In this paper, we also propose to utilize head shuffling to further improve the performance of the Transformer structure as shown in Fig.~\ref{fig:shuffle-txr}. In the original Transformer~\cite{vaswani2017attention}, the feature vectors of all heads are directly concatenated before being fed into the last fully-connected layer. Thus, there is no interaction between those heads from the start to the end. To enable the interactions across heads, we propose to divide each head and shuffle all head vectors before passing them on to separate fully-connected layers. The outputs are finally concatenated in a late fusion style. This scheme is similar to ShuffleNet~\cite{zhang2018shufflenet}, with the key difference that here we conduct shuffling between different heads within the multi-head attention, while in ShuffleNet the shuffling is across channels. Our empirical results show that our shuffling operation results in better generalization of the model.

\noindent \textbf{Decoder:} For the generation setting, with the final encoded feature $h^{\star}_\text{enc}$, we use a feed-forward network with softmax to predict the next token probability distribution $P$ over all tokens in the vocabulary $\mathcal{V}$; i.e., $P = \mathrm{softmax}(\mathrm{FFN}(h^{\star}_\text{enc}))$.
In the testing stage, we conduct beam search with $b$ beams to generate an answer sentence.

\noindent \textbf{Loss Function:} Let $\mathcal{P}$ denote the collection of all next-token probability distributions $P_j\in\mathbb{R}^{|\mathcal{V}|}$, $j=1,\dots,N$ for batch size $N$, and let $\mathcal{G}$ be the collection of respective distributions $G_j$ for the ground truth answer tokens. For the generation setting, we apply label smoothing~\cite{muller2019does} to account for the sparsity of the token distributions, leading to $\tilde{\mathcal{G}}$. 
We use the cross-entropy ($\mathrm{CE}$) loss between the predicted and the smoothed ground truth distributions to train our model end-to-end:
\begin{equation} 
\mathcal{L} = \mathrm{CE}(\mathcal{P} | \tilde{\mathcal{G}}) = -\frac{1}{N}\sum_{j=1}^{N} \sum_{u\in \mathcal{V}} \tilde{G}_j(u)\log P_j(u).
\label{eq:loss}
\end{equation}
For the retrieval setting, we first concatenate the feature embeddings of the query and the various input modalities obtained from the Encoder module of our network ($e_h, e_c, e_q, e_v$). Next, the candidate answers are embedded into this joint space using LSTMs, and a dot product is taken between the concatenated inputs and embeddings of each of the answer candidates. We then train this model with the binary cross-entropy loss.
\section{Experiments}
In this section, we detail our experimental setup, datasets, and evaluation protocols, before furnishing our results.
\noindent\textbf{Dataset and Evaluation:} We use the audio-visual scene-aware dialog (AVSD) dataset~\cite{alamri2019audio} for our experiments, which is the benchmark dataset for this task. This dataset emulates a real-world human-human natural conversation scenario about an audio-visual clip. See~\cite{alamri2019audio} for details of this task and the dataset. We evaluate on two variants of this dataset corresponding to annotations available for the DSTC-7 and DSTC-8 challenges,\footnote{\url{https://sites.google.com/dstc.community/dstc8/home}} consisting of 7,659, 1,787, 1,710, and 1,710 dialogs for training, validation, DSTC-7 testing, and DSTC-8 testing, respectively for the answer generation task. The quality of the generated answers is evaluated using the standard MS COCO evaluation metrics~\cite{chen2015microsoft}, such as BLEU, METEOR, ROUGE-L, and CIDEr. Apart from the answer generation task~\cite{hori2018multimodal}, we also report experiments on the answer selection task, described in~\cite{alamri2019audio} using their annotations and ground truth answers. This task requires selecting the answer to a question from a set of 100 answers. Specifically, in this task, an algorithm is to present a ranking over a set of 100 provided answers, with ideally the correct answer ranked as the first. The evaluation is then based on the mean retrieval rank over the test set.

\noindent \textbf{Data Processing:}
We follow~\cite{le2019multimodal} to perform text preprocessing which include lowercasing, tokenization, and building a vocabulary by only selecting tokens that occur at least five times. Thus, we use a vocabulary with 3,254 words, both for the generation and retrieval tasks.

\noindent \textbf{Feature Extraction:}
Motivated by ~\cite{anderson2018bottom}, we train a detector on Visual Genome with 1601 classes and 401 attributes, which incorporates a ``background'' label and a ``no-attribute'' label. We use ResNext-101 as the neural backbone with a multiscale feature pyramid network. We further use fine-grained ROI-alignment instead of ROI-pooling for better feature representation. We extract the 1024-D features for the 36 highest scoring regions, their class labels, and attributes. After extracting the region features, we apply a pretrained relationship detector~\cite{zhang2019vrd} to find visually-related regions. We calculate the minimal bounding box which covers two visually-related regions and perform ROI-alignment to get compact representations for relationship regions. 
In order to incorporate audio into the STSGR framework, we extract AudioSet VGG-ish features~\cite{hershey2017cnn} from the audio stream for every video. These are 128-D features obtained from the AudioSet VGG-ish CNN, pretrained on 0.96s Mel Spectrogram patches on the AudioSet data~\cite{gemmeke2017audio}. 

\begin{table}[t]
\centering
\resizebox{0.435\textwidth}{!}{
\begin{tabular}{l|cccc}
    \hline
    Method & B4 & MET & ROUGE  & CIDEr \\
    \hline
     STSGR full model &  \textbf{0.133} & \textbf{0.165} & \textbf{0.361} & \textbf{1.265}\\ 
     STSGR w/o shuffle   & 0.127 & 0.161 & 0.354 & 1.208 \\
     STSGR w/o GAT   & 0.118 & 0.160 & 0.347 & 1.125 \\
     STSGR w/o EdgeConv   & 0.131 & 0.162 & 0.356 & 1.244 \\
     STSGR w/o union box features  & 0.124 & 0.163 & 0.352 & 1.175 \\
     STSGR w/o visual features & 0.127 & 0.160 & 0.356 & 1.203 \\
     STSGR w/o temporal & 0.125 & 0.164 & 0.357 & 1.212 \\
     \hline
     STSGR + audio  & \textbf{0.133} & \textbf{0.165} & \textbf{0.362} & \textbf{1.272}\\
     \hline
\end{tabular}
}
\caption{Ablation study using AVSD@DSTC7 dataset.}
\label{tab:ablation}
\vspace{-0.6cm}
\end{table}

\noindent \textbf{Model Training:}
We set our Transformer hyperparameters following \cite{vaswani2017attention}. The feature dimension is 512, while the inner-layer dimension of the feed-forward network is set to 2048. For multi-head attention, we maintain $h=8$ parallel attention heads and apply shuffling to boost performance. For the semantic labels, we build a 300-D embedding layer for the 1651 words in the vocabulary (which is available with the dataset), and initialize the embeddings using GloVe word vectors~\cite{pennington2014glove}. For semantic labels consisting of more than one word, we use the average word embedding as the label embedding. Our model is trained on one Nvidia Titan XP GPU with Adam optimizer~\cite{kingma2015adam} with $\beta_1 = 0.9$, $\beta_2 = 0.98$. The batch size is set to 16 and we adopt the warm-up strategy as suggested in~\cite{vaswani2017attention} for learning rate adjustment with about 10,000 steps.

\noindent \textbf{Baselines:} We consider the following four baselines on the generation task: (i) \textit{Baseline}~\cite{hori2019end}, (ii) \textit{Multi-modal Attention}~\cite{hori2019end}, that uses attention over concatenated features, (iii) \textit{Simple}~\cite{schwartz2019factor} that uses factor-graph attention on the modalities, and (iv) \textit{MTN}~\cite{le2019multimodal} that applies self-attention and co-attention to aggregate multi-modal information. For the retrieval task, we compare our method against the state-of-the-art method of ~\cite{alamri2019audio} on the DSTC-7 split.

\begin{figure*}[ht]
  \centering
  \includegraphics[width=16.9cm]{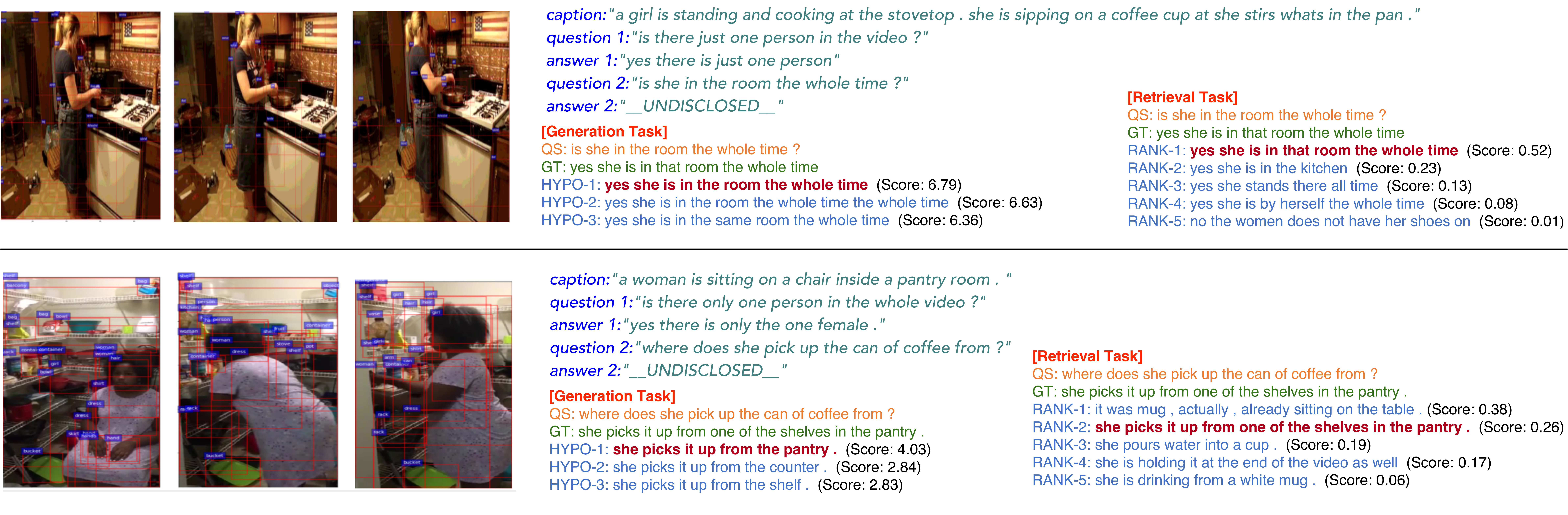} 
\caption{Qualitative results from our model on both generation and retrieval tasks of AVSD. Left: input video frames, Top-right: caption and dialog history, Bottom-middle: top-3 generated answers with confidence scores. Bottom-right: top-5 ranked candidate answers with confidence scores.}
\vspace{-.35cm}
\label{fig:quals}
\end{figure*}

\noindent\textbf{Ablation Study:} To understand the importance of each component in our model, Table~\ref{tab:ablation} details an ablation study. We analyze several key components: (i) shuffling in the Transformer structure, (ii) visual and semantic graph, (iii)  ROI Recrop on the union bounding boxes, and (iv) temporal aggregation. From the table, we see that Graph Attention Network (GAT), which is used to produce the visual scene graph, is important to aggregate information from neighboring nodes (e.g., improving CIDEr from 1.125 to 1.265), while EdgeConv, used in the semantic graph, offers some improvement (e.g., CIDEr from 1.244 to 1.265). Moreover, the use of shuffling in the multi-head Transformer architecture boosts the performance significantly (from 1.208 to 1.265 for CIDEr). We can also conclude that union bounding boxes, semantic labels, and inter-frame aggregation contribute to stabilize the generation performance. Overall, by adopting all these key components, the full model outperforms all the ablations. From Tables~\ref{tab:ablation} and ~\ref{tab:audio_ret}, we notice that incorporation of audio helps improve the performance of our model. For instance, on the retrieval setting we observe that incorporating audio lowers the Mean-Retrieval Rank noticeably down to 4.08 from 4.33 for the full model and to 5.91 from 6.54 when no language context is available.

\begin{table}[t]
\centering
\resizebox{0.5\textwidth}{!}{
\begin{tabular}{l|cccc}
\hline
\multicolumn{5}{c}{AVSD@DSTC7}\\
\hline
Method &  B4 & MET & ROUGE & CIDEr\\
\hline
    Baseline~\cite{hori2019end}  & 0.075 & 0.110 & 0.275 & 0.701   \\
     Multi-modal Attention~\cite{hori2019end}  &  0.078 & 0.113 & 0.277 & 0.727 \\
     Simple~\cite{schwartz2019simple}  &  0.091 & 0.125 & 0.307 & 0.883\\
     MTN~\cite{le2019multimodal}  &  0.128 & 0.162 & 0.355 & 1.249\\
     Ours & \textbf{0.133} & \textbf{0.165} & \textbf{0.362} & \textbf{1.272}\\
     \hline
     \multicolumn{5}{c}{AVSD@DSTC8}\\
     \hline
     Baseline~\cite{hori2019end} & 0.289 & 0.210 & 0.480 & 0.651 \\
     Multi-modal Attention~\cite{hori2019end}  &  0.293 & 0.212 & 0.483 & 0.679 \\
     Simple~\cite{schwartz2019simple} & 0.311 & 0.224 & 0.502 & 0.766\\
     MTN~\cite{le2019multimodal}  &  0.352 & 0.263 & 0.547 & 0.978\\
     Ours & \textbf{0.357} & \textbf{0.267} & \textbf{0.553} & \textbf{1.004}  \\
\hline
\end{tabular}
}
\caption{Comparisons of our method against the state of the art on the AVSD test splits for DSTC7 and DSTC8.}
\label{tab:dstc7}
\vspace*{-.2cm}
\end{table}

\begin{table}[t!]
\centering
\resizebox{0.469\textwidth}{!}{
\begin{tabular}{l|ccc}
\hline
Method &  Full model & w/o caption & w/o cap. Diag. Hist.\\
\hline 
     \citet{alamri2019audio} & 
     5.88 & N/A & 7.41 \\
     \citet{hori2019end} & 
     5.60 & N/A & 7.23 \\
     MTN w/o audio & 
     4.51 & 4.90 & 6.85 \\ 
     MTN w/ audio & 
     4.29 & 4.78 & 6.46 \\ 
     STSGR & 
     4.33 & 4.67 & 6.54 \\ 
     STSGR w/ audio & 
     \textbf{4.08} & \textbf{4.55} & \textbf{5.91} \\ 
\hline
\end{tabular}
}
\caption{State-of-the-art comparisons on answer selection as measured by Mean Retrieval Rank (lower the better).}
\label{tab:audio_ret}
\vspace*{-.15cm}
\end{table}

\noindent\textbf{Comparisons to the State of the Art:} In Table~\ref{tab:dstc7}, we compare STSGR against baseline methods on various quality metrics based on ground-truth answers.  As is clear, our approach achieves better performance against all the baselines. The performance  on the answer selection task (mean retrieval rank) is provided in Table~\ref{tab:audio_ret}, demonstrating clearly state-of-the-art results against the baseline in~\cite{alamri2019audio}. We also show that including audio into the STSGR representation helps improve the mean retrieval rank. 

\begin{table}[t!]
\centering
\resizebox{0.43\textwidth}{!}{
\begin{tabular}{c|c|cccc}
\hline
Method & Feature & B4 & M & R & C \\
\hline 
     Simple & i3d &
     0.091 & 0.125 & 0.307 & 0.883 \\
     Simple & VGG &
     0.095 & 0.126 & 0.309 & 0.919 \\
     MTN & N/A &
     0.114 & 0.147 & 0.332 & 1.106 \\
     MTN & i3d &
     0.118 & 0.160 & 0.348 & 1.151 \\
     STSGR (Ours) & N/A &
     0.121 & 0.152 & 0.350 & 1.186 \\
     STSGR (Ours) & i3d &
     0.122	& 0.152 & 0.353	& 1.223 \\
     STSGR (Ours) & Scene Graphs &
     \textbf{0.133} & \textbf{0.165} & \textbf{0.361} & \textbf{1.265} \\
\hline
\end{tabular}
}
\caption{Comparison of visual representations on DSTC7.}
\label{tab:avsd}
\vspace*{-.5cm}
\end{table}

\noindent\textbf{Qualitative Results and Discussion:}  In Fig.~\ref{fig:quals}, we provide two qualitative results from our STSGR model. For the first case, our model consistently detects the woman in the frames and finds that she maintains many connections with other objects inside the scene throughout the whole video, thus our model makes/selects the correct answer with high confidence. For the second case, the clutter background poses a challenge to our model. However, STSGR can still generate/rank the correct answer in top-2. 
In general, we find that STSGR can answer spatial and temporal questions very well. This is quantitatively evidenced by observing that while both STSGR and MTN~\cite{le2019multimodal} use similar backends, they differ in the input representations (I3D in~\cite{le2019multimodal} vs. scene graphs in ours), and our model outperforms MTN noticeably (1.272 vs 1.249 on CIDEr, Table~\ref{tab:dstc7}), substantiating the importance of our STSGR representation. In Table~\ref{tab:avsd}, we further compare STSGR representation against other visual representations (I3D, VGG) and different methods (Simple~\cite{schwartz2019simple}, MTN~\cite{le2019multimodal} on the generation task, and demonstrate that our proposed scene graph representation by itself is a better way to characterize visual content.
\section{Conclusions}
We presented a novel hierarchical graph representation learning and Transformer reasoning framework for the problem of audio-visual scene-aware dialog. Specifically, our model generates object, frame, and video-level representations that are systematically integrated to produce visual memories, which are sequentially fused to the encodings of other modalities (dialog history, etc.) conditioned on the input question using a multi-head shuffled Transformer. Experiments demonstrate the benefits of our framework for both generation/selection tasks on the AVSD benchmark. Going forward, we plan to explore the use of richer text embeddings (such as GPT~\cite{radford2019language} and BERT~\cite{devlin2018bert}) within our framework. 

\noindent{\textbf{Ethical Impact:}}
Our proposed STSGR framework presents a step towards improved modelling of multimodal data and we believe it could be a valuable addition to artificial intelligence research. However, STSGR does not explicitly account for biases in the training set, and thus may result in the model capturing  social prejudices.

\noindent\textbf{Acknowledgements:} Shijie Geng, Peng Gao, and Moitreya Chatterjee worked on this project during MERL internships.

\bibliography{stsgr.bib}
\end{document}